\documentclass[conference]{IEEEtran}
\IEEEoverridecommandlockouts
\usepackage{cite}
\usepackage{amsmath,amssymb,amsfonts}
\usepackage{algorithmic}
\usepackage{graphicx}
\usepackage{caption}
\usepackage{subcaption}
\usepackage{textcomp}
\usepackage{xcolor}
\usepackage{hhline}
\usepackage{comment}
\usepackage{multirow,multicol}
\def\BibTeX{{\rm B\kern-.05em{\sc i\kern-.025em b}\kern-.08em
    T\kern-.1667em\lower.7ex\hbox{E}\kern-.125emX}}
\begin{document}

\title{Investigation of BERT Model on Biomedical Relation Extraction Based on Revised Fine-tuning Mechanism}

\author{\IEEEauthorblockN{ Peng Su}
\IEEEauthorblockA{\textit{Department of Computer and Information Science} \\
\textit{University of Delaware}\\
Newark, Delaware, USA \\
psu@udel.edu}
\and
\IEEEauthorblockN{K. Vijay-Shanker}
\IEEEauthorblockA{\textit{Department of Computer and Information Science} \\
\textit{University of Delaware}\\
Newark, Delaware, USA\\
vijay@udel.edu}

}

\maketitle

\begin{abstract}
With the explosive growth of biomedical literature, designing automatic tools to extract information from the literature has great significance in biomedical research. Recently, transformer-based BERT models adapted to the biomedical domain have produced leading results. However, all the existing BERT models for relation classification only utilize partial knowledge from the last layer. In this paper, we will investigate the method of utilizing the entire layer in the fine-tuning process of BERT model. To the best of our knowledge, we are the first to explore this method. The experimental results illustrate that our method improves the BERT model performance and outperforms the state-of-the-art methods on three benchmark datasets for different relation extraction tasks. In addition, further analysis indicates that the key knowledge about the relations can be learned from the last layer of BERT model.  
\end{abstract}


\section{Introduction}
In the past few decades, biomedical literature has been growing explosively and its interdisciplinary nature makes biomedical text mining a unique part of modern biomedical research. Among text mining tasks, relation extraction (RE) plays a crucial role in extracting the structure of the text. The task of relation extraction in biomedical domain is to identify semantic relationships between biomedical entities mentioned in the text. For example, protein-protein interaction is a RE task that involves the extraction of the interacting proteins from the literature. RE allows information available in unstructured form in the literature to be stored in a structured form that can be used in many downstream applications like database curation, question answering system, text summarization etc.

Many different methods have been proposed to solve the RE problem \cite{sierra2008definitional,culotta2004dependency,zhang2019deep,sahu2018drug}. Recently, utilizing neural pre-trained language models has been shown to be an effective way to improve model performance of natural language processing (NLP) tasks including RE \cite{dai2015semi, peters2018deep, devlin2018bert, radford2019language}. These methods can make best use of a large amount of unlabeled data in an unsupervised pre-training phase to build language representation. This phase can be followed by supervised fine-tuning on task-specific datasets. In this work, we will employ the BERT \cite{devlin2018bert}, a language representation model based on bidirectional Transformer \cite{vaswani2017attention}, to solve the relation extraction problem. In the pre-training stage, BERT learns general language representation for both the words and the text sequence through two unsupervised tasks. This task-agnostic representation is applied for the specific tasks via supervised learning on task-specific datasets in the fine-tuning stage.

Since the development of BERT \cite{devlin2018bert}, previous works on relation extraction and other classification tasks have all utilized the suggested approach of using a special token called the classification token (CLS) in the topmost layer. Thus, during fine-tuning for classification tasks, all information is funneled through a part of the top layer during back-propagation. In this work, we will explore the potential of utilizing all the information in the last layer to improve model performance on relation extraction tasks.

Our work is partly driven by an observation in  \cite{tenney2019bert}, where the nature of information represented in various layers was studied. The authors found that BERT models appeared to mimic traditional NLP approaches where low-level morphological and syntactic aspects are represented in lower layers and higher level and more semantic nature of text is learned in higher layers of BERT model. Since relation extraction task is clearly semantic in nature, we hypothesize that important information for this task will reside in top layers of the BERT model. We therefore wish to investigate whether all the information in the top layer of the BERT model could be fruitfully used to improve the performance of RE models.

To test our hypothesis, we employ two different methods to summarize the information in the last layer: Long Short Term Memory (LSTM) model \cite{hochreiter1997long} and attention mechanism \cite{bahdanau2014neural}, and then include the summarized knowledge in a new fine-tuning process. We test these methods on three widely studied RE tasks in biomedical domain: the protein-protein interaction (PPI) \cite{krallinger2008overview}, the drug-drug interaction (DDI) \cite{herrero2013ddi}, and the chemical-protein interaction (ChemProt) \cite{krallinger2017overview}. Our results show that utilizing all the information from the last layer boosts the model performance on all three tasks and furthermore, we now achieve the state-of-the-art results on the three benchmark datasets.

The rest of this manuscript is organized as follows. We discuss some related works and background in Section 2. In Section 3, we describe the details of methodology and experimental settings, followed by the evaluation results and discussion in Section 4. We conclude in the last section.

\section{Related Work and background}

We first present some related works in this section. Then we briefly introduce BERT model, its pre-training and fine-tuning process.

\subsection{Related Work}

Extracting various of relations in BioNLP domain has attracted considerable attentions in recent years with deep learning methods and high-quality word representation (word embedding) showing advancements in performance \cite{mikolov2013distributed, pennington2014glove, chiu2016train}. Both convolutional neural networks (CNN) and recurrent neural networks (RNN) \cite{hsieh2017identifying, zhang2019deep,su2019using} have achieved notable results on relation extraction tasks based on word embedding in this field before the emerging of utilizing language model. 

An advantage of language model methods is that they can leverage a large amount of unlabeled data to learn the context-dependent and universal language representation. The pre-trained language model can be applied on downstream tasks through fine-tuning and many language models have been proposed such as ELMo \cite{peters2018deep}, GPT \cite{radford2018improving} and BERT \cite{devlin2018bert}. Peters et al. proposed a contextualized word representation from a pre-trained language model (ELMo) and significantly improved the performance on six tasks of question answering, textual entailment and sentiment analysis \cite{peters2018deep}. The Generative Pre-trained Transformer (GPT) \cite{radford2018improving} was based on a unidirectional language model and has shown its effectiveness on benchmarks for natural language understanding. BERT (Bidirectional Encoder Representations from Transformers) model \cite{devlin2018bert} was designed to learn language representation in bidirectional way and was demonstrated to obtain state-of-the-art performance on tasks such as question answering and language inference. 

Usually, pre-trained BERT model for general domain generalizes poorly in a specific domain like biomedical domain. In order to facilitate the biomedical research, several BERT models have been adapted for the biomedical domain such as BioBERT \cite{lee2020biobert}, SciBERT \cite{beltagy2019scibert}, BlueBERT \cite{peng2019transfer}. These models are pre-trained on different types of biomedical corpora and have been shown to achieve state-of-the-art results on various biomedical tasks. BioBERT \cite{lee2020biobert} was pre-trained on PubMed abstracts and PMC full-length articles, BlueBERT \cite{peng2019transfer} used PubMed abstract and MIMIC-III clinical notes \cite{johnson2016mimic} as the pre-training data, and a sample of 1.14M papers from Semantic Scholar \cite{ammar2018construction} were used in SciBERT model \cite{beltagy2019scibert}. In this work, we will first select the best BERT model on our relation extraction tasks and then employ that model in the following experiments.

However, most of previous works only use the output of classification token ([CLS]) in BERT model for classification problem and ignore other outputs, which contains useful information. In the work \cite{song2020utilizing}, the authors explored the methods of extracting knowledge from the intermediate layers in BERT and demonstrated the advancement on aspect based sentiment analysis and natural language inference tasks. Since residual connections are employed in the Transformer, all important information in intermediate layers should be transmitted to the last layer of BERT model during training. We will compare this method with our approach of utilizing all the information from the last layer in BERT model. 

In addition, the work \cite{tenney2019bert} found that the BERT model represents the steps of the traditional NLP pipeline in a sequential way, which means the BERT model captures the linguistic information in the sequence of part-of-speech (POS) tagging, parsing, NER (Named Entity Recognition), semantic roles, then coreference. Their experiments also showed that the representation for relation classification continues to improve up to the highest layer of the BERT model, indicating that the last layer contains important information for the relation extraction tasks.

Our method evaluation will be conducted on three relation extraction tasks: PPI \cite{krallinger2008overview}, DDI \cite{herrero2013ddi}, and ChemProt \cite{krallinger2017overview}. For PPI, it is probably the most widely studied relation extraction task in biomedical domain because its key role in understanding biological processes. Extracting relation of DDI will help prevent adverse effects from drug combinations. ChemProt relation extraction will benefit precision medicine by finding the affected proteins by a specific chemical. Dozens of systems have been designed for these tasks like rule/kernel-based methods \cite{airola2008all, kim2015extracting, warikoo1252017ctcpi}, neural network methods \cite{peng2017deep, zhang2019deep, liu2016drug}, language representation model methods \cite{lee2020biobert, peng2019transfer}. As far as we know, the state-of-the-art system on AIMed corpus for PPI task is based on deep residual convolutional neural network \cite{zhang2019deep} while the best results on DDI and ChemProt tasks are obtained using the BERT model \cite{lee2020biobert, peng2019transfer}.

\subsection{Background}
BERT \cite{devlin2018bert} is a bidirectional Transformer model \cite{vaswani2017attention} for language representation. Using a "masked language model", BERT is able to learn bidirectional representation (both left and right context) of a word. The application of BERT usually includes two steps: utilizing pre-training to gain general knowledge and employing fine-tuning to acquire task-specific knowledge of a task. 

\textbf{Pre-training of BERT}: BERT is pre-trained using two unsupervised tasks: masked language model (MLM) and next sentence prediction (NSP). The pre-training procedure usually involves a large amount of unlabeled data. Originally, BERT was designed for general purpose and it was pretrained on English Wikipedia and BooksCorpus. However, each domain has its specific knowledge and sometime it is impossible for the general BERT to gain such knowledge. For example, in biomedical domain, the biomedical entity names can not be represented well based on the general BERT model since its pre-training corpora are short of such names. In this paper, we will consider three BERT models that are adapted to the biomedical domain (BioBERT \cite{lee2020biobert}, SciBERT \cite{beltagy2019scibert} and BlueBERT \cite{peng2019transfer}) and select the best one for the experiments on the new fine-tuning process.  

\begin{figure}[tb]
\centering
\includegraphics[width=0.45\textwidth]{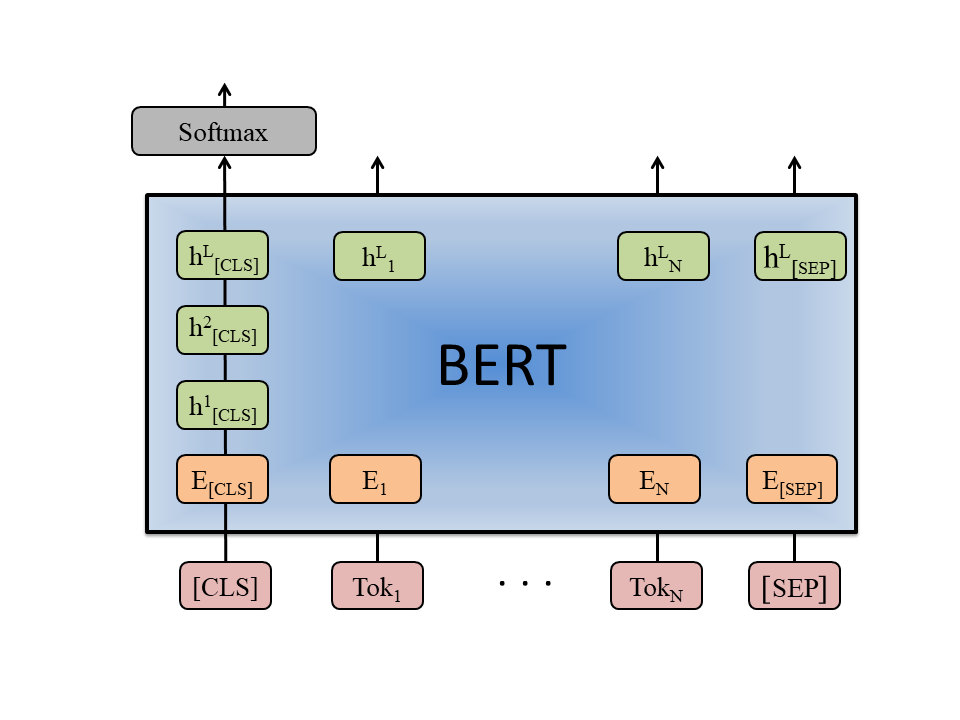}
\caption{\label{fig:bert} Original BERT model architecture on relation extraction.}
\end{figure}

\textbf{Fine-tuning of BERT}: The pre-trained BERT model can be fine-tuned for various downstream tasks. Given a task, we just need to append an additional output layer on top of the BERT output and then feed the task-specific inputs and outputs into BERT model and fine-tune the BERT parameters like regular model training. For classification problems, a classifier can be built by putting a Softmax layer on the [CLS] token output of pre-trained BERT model. The [CLS] token is designed for classification purpose in BERT model and every input sequence will always start with this special token. The model architecture of fine-tuning based on [CLS] token is shown in Fig \ref{fig:bert}.

\section{Methodology}
In this section, we first define the relation extraction problem. Then, we propose the method of adding a new module to summarize the output information in the last layer during fine-tuning. The evaluation datasets and metrics, the fine-tuning parameters and pre-processing of data for BERT model are also given in this section.

\subsection{Relation extraction}
Relation extraction task can be seen as a classification problem. For example, for the DDI relation extraction task, a sentence instance with two marked drug mentions is considered one of the four relations (ADVICE, EFFECT, INT, MECHANISM) if the sentence expresses drug interaction between the two drug occurrences, otherwise it will be seen as negative. Formally, for a relation $R$ and its label set $L$, our model needs to predict the probability of each label $P(L|Tok_1,\,Tok_2,\,\ldots,Tok_n, Et_1, Et_2)$ based on the sentence text $S=Tok_1 \,Tok_2 \, \ldots Tok_n$ and two entities $Et_1$, $Et_2$ in the sentence. BERT model can be applied on this relation extraction classification problem.

\begin{figure*}
     \begin{subfigure}[b]{0.46\textwidth}
         \centering
         \includegraphics[width=\textwidth]{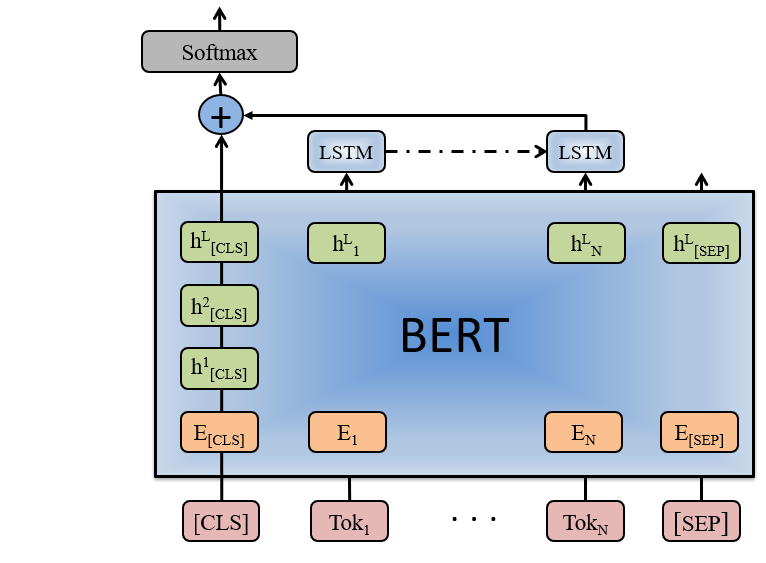}
         \caption{LSTM on the last layer}
         \label{fig:modelc}
     \end{subfigure}
     \hfill
     \begin{subfigure}[b]{0.47\textwidth}
         \centering
         \includegraphics[width=\textwidth]{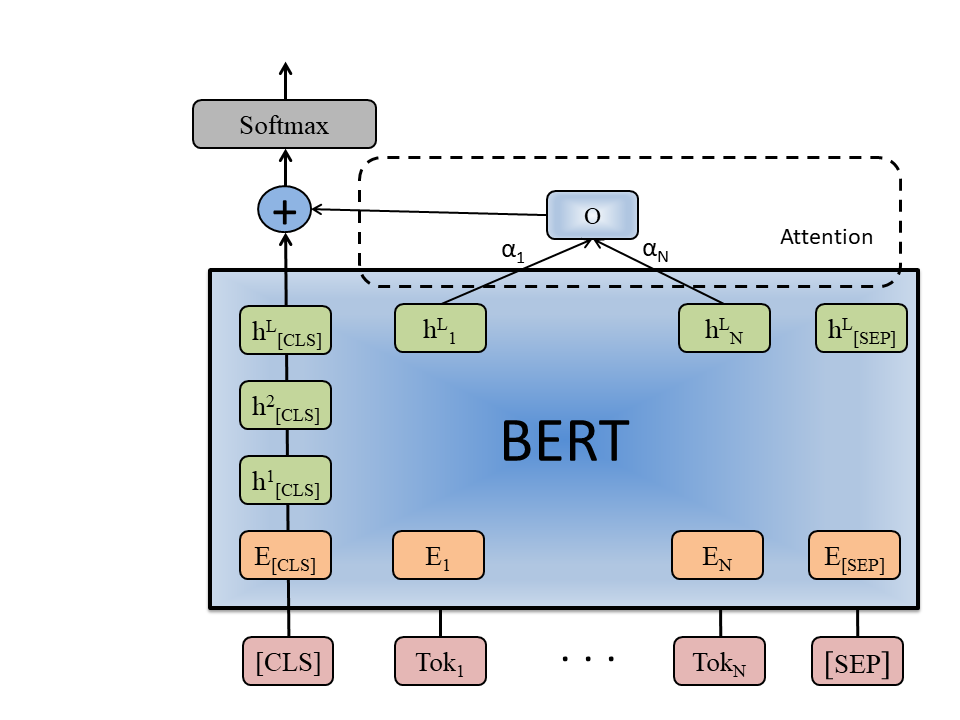}
         \caption{Attention mechanism on the last layer}
         \label{fig:modeld}
     \end{subfigure}
        \caption{Model architectures of including all outputs in the last layer.}
        \label{fig:model}
\end{figure*}

\subsection{New fine-tuning process of BERT model}

As mentioned above, a BERT model for classification problem is usually based on the [CLS] token of the last layer and other outputs of the last layer are abandoned. So this method might discard some useful information from the last layer. The [CLS] token is used to predict the next sentence (NSP task) during the pre-training, which usually involves two or more sentences, but the input of our relation extraction tasks only contain one sentence. This indicates that the [CLS] output might ignore important information about the entities and their interaction because it is not trained to capture this kind of information. The fine-tuning step on task-specific datasets compensates this information loss on relations to a certain extent, but utilizing all information might yield a better representation than the use of partial information (only the [CLS] output). As it is also shown in \cite{tenney2019bert}, higher layer outputs yield better representations for the relation classification task. In this paper, we will design a mechanism to utilize all the information in the last layer during the fine-tuning of BERT model.    

To include the information from other outputs of the last layer, we propose to add an new module to summarize these outputs and then concatenate the summarized information with the [CLS] output as the extra step for fine-tuning. Specifically, we explore two methods of summarizing the information in the last layer: Long Short Term Memory (LSTM) model and attention mechanism. The model architectures of these two methods are illustrated in Fig \ref{fig:model}. 

Formally, given the hidden states (with dimension of H) of the last layer (L-th layer) $$h_{all}^L=\{h_{CLS}^L,\ h_1^L,\ h_2^L,\ ...,\ h_{N}^L,\ h_{SEP}^L\},$$ only $h_{CLS}^L$ is used for classification in the original BERT model. Here we will try to add the summarized information from the sequence of other outputs, a.k.a., $h^L=\{h_1^L,\ h_2^L,\ ...,\ h_{N}^L\}$ (We ignore $h_{SEP}^L$ since it is a separation token for different sentences) and two different approaches are employed: LSTM sequence model and attention mechanism.
\begin{enumerate}
 \item A natural idea to model the sequence of the last layer is to use recurrent neural network and LSTM model is our first choice since its advantages on long sequence. The output of last LSTM cell will be seen as the representation of our sequence. Formally, $$O=LSTM(h_i^L)\quad i=1,...,N$$
 \item Another choice of summarizing sequence data is the attention mechanism. Specifically, we can utilize additive attention to combine the sequence by assigning a weight for each output. $$[\alpha _i] = softmax(h^LK)$$ $$O = \sum^N_{i=1}\alpha_i h_i^L$$ where $K^{H \times 1}$ are learnable parameters.
\end{enumerate}

We then concatenate the output of LSTM/attention mechanism with the [CLS] output to form the final representation of our input $$h =[h_{CLS}^L; O].$$ Finally, we feed this representation to a softmax layer to predict the label: $$p=softmax(W_fh+b_f)$$ where $W_f^{C \times H}$, $b_f^{C \times 1}$ are also learnable parameters and $C$ is the number of classes (categories) in the classification task.

Meanwhile, the contribution of [CLS] token for the classification task needs to be explored. So we experiment with another fine-tuning process without [CLS] token. Formally, we only take $h =[O]$ as the final representation of the input and utilize the same softmax layer $p=softmax(W_fh+b_f)$ for prediction. The performance of these two new fine-tuning processes can shed light on the roles of the two parts (the [CLS] token and sentence outputs) of the last layer for classification problem.   

\begin{table}
\centering
\begin{tabular}{ c c c c c }
\hline

\hline

\hline

\hline
Corpus &  Instance\# (Total) & Train & Dev & Test \\
\hline
PPI(AIMed) & 5,834 & - & - &- \\
DDI & 48,223 & 19,460 & 11,820 & 16,943 \\
ChemProt & 31,784 & 18,779 & 7,244 & 5,761 \\
\hline

\hline

\hline

\hline
\end{tabular}
\caption{Statistics of datasets for PPI, DDI and ChemProt.}
\label{tab:statistics}
\end{table}

\begin{table*}[tb]
\centering
\begin{tabular}{c|l c }
\hline

\hline

\hline

\hline
Task & Label&Sentence examples\\
 \hline
\multirow{2}{3.8em}{PPI} &Positive  & We have characterized the physical and functional interactions between @PROTEIN\$ and @PROTEIN\$.\\
&Negative & pRB, @PROTEIN\$, @PROTEIN\$(INK4d) and p27(KIP1) decrease in both cell types.\\
 \hline
\multirow{2}{3.8em}{DDI}&MECHANISM  & @DRUG\$ decreases the elimination of @DRUG\$ causing an increase in overall exposure. \\
&EFFECT & Anticoagulants @DRUG\$ may increase sensitivity to oral @DRUG\$.\\
\hline
\multirow{3}{3.8em}{ChemProt}&CPR:4 & @CHEMICAL\$ potently attenuated gene expressions involved in inflammation, such as iNOS, COX-2 and @GENE\$.\\
&\multirow{2}{3.8em}{CPR:9} & They suggest that TRPML1 works in concert with @CHEMICAL\$ to regulate @GENE\$ translocation between \\&&the cytoplasm and lysosomes. \\
\hline

\hline

\hline

\hline

\end{tabular}
\captionsetup{width=0.99\textwidth}
\caption{Examples after pre-processing from the three tasks.}
\label{tab:sents}
\end{table*}

\subsection{Datasets and evaluation metrics}

To measure the effectiveness of the new fine-tuning mechanism, we will evaluate it on three benchmark datasets. Table~\ref{tab:statistics} shows the statistics of these datasets. For PPI task, we will utilize the most well-known PPI relation corpus: AIMed \cite{bunescu2005comparative}. It is obtained by annotating 750 abstracts from Medline and it contains 1,000 positive (interacting protein pairs) and 4,834 negative (non-interacting protein pairs) instances. Since there is no standard training and test set for this dataset, we will employ 10-fold cross-validation on it. 

The corpus for ChemProt task contains a training set of 1,020 abstracts, a development set of 612 abstracts and a test set of 800 abstracts from PubMed \cite{krallinger2017overview}.  We will use the standard training and test set to evaluate the model performance. 

For the DDI corpus, it has 792 documents from DrugBank database and 233 abstracts from Medline \cite{herrero2013ddi}. Similar to ChemProt, we will use the standard split of this dataset for evaluation (624 documents for training, 210 for development and 191 for test). 

As for the evaluation metrics, PPI is a binary classification problem, so we will use precision, recall and F1-score to measure the performance of the models. However, the ChemProt and DDI tasks are multi-class classification problem. The ChemProt corpus is labeled with six classes: CPR:3 (upregulator, activator, indirect\_upregulator), CPR:4 (downregulator, inhibitor, indirect\_downregulator), CPR:5 (agonist, agonist-activator, agonist-inhibitor), CPR:6 (antagonist), CPR:9 (substrate, product\_of, substrate\_product\_of) and negative. In DDI corpus, there are five labels for the instances: ADVICE, EFFECT, INT, MECHANISM and negative. For ChemProt and DDI tasks, we evaluate the model utilizing micro (i.e., with weighted mean) precision, recall and F1 score on the non-negative classes as other people report their results in the same way.
\begin{table*}
\centering
\begin{tabular}{c c c c | c c c | c c c }
\hline

\hline

\hline

\hline
\multirow{2}{4em}{Model}&\multicolumn{3}{c|}{PPI}&\multicolumn{3}{c|}{DDI} & \multicolumn{3}{c}{ChemProt}\\
\cline{2-10}
&  Precision & Recall & F1 Score & Precision & Recall & F1 Score & Precision & Recall & F1 Score \\
\hline
SOTA & 79.0 & 76.8& 77.6& 79.3 & \textbf{80.5} & 79.9& \textbf{77.0} & 75.9 & 76.5  \\
BioBERT &79.0 & 83.3 & 81.0 &77.4 & 79.4 & 78.4 & 74.6 & 76.8 & 75.7\\
\hline
\hline
BioBERT+LSTM\_LL &80.2 & 84.0 & 82.0 &79.4 & 78.0 & 78.7 & 75.7 & \textbf{77.4} & 76.5\\
BioBERT+Att\_LL &\textbf{80.7} & \textbf{84.4} & \textbf{82.5} &\textbf{81.3} & 80.1 & \textbf{80.7} & 76.5 & 77.1 & \textbf{76.8}\\
\hline

\hline

\hline

\hline
\end{tabular}
\captionsetup{width=.85\textwidth}
\caption{Model performance on PPI, DDI and ChemProt tasks. SOTA: state-of-the-art results for each task; BioBERT+LSTM\_LL: model of applying LSTM on the outputs of last layer; BioBERT+Att\_LL: model of applying attention mechanism on the outputs of last layer. To the best of our knowledge, the state-of-the-art systems on the three tasks are: \cite{zhang2019deep} for PPI; \cite{lee2020biobert} for ChemProt; \cite{peng2019transfer} for DDI.}
\label{tab:comp}
\end{table*}

\subsection{Fine-tuning hyper-parameters}

During the fine-tuning of the models, we use learning rate of 2e-5, batch size of 32, training epoch of 10 and max sequence length of 128. 

\subsection{Data pre-processing}

In relation extraction tasks, the input for the BERT model contains two parts: the text and the entities in it. In order for BERT model to recognize the entities in the sentence, we follow the standard pre-processing way for relation extraction by replacing these entity names with predefined tags. Specifically, we replace all the protein names with @PROTEIN\$, drug names with @DRUG\$, and chemical names with @CHEMICAL\$. In Table \ref{tab:sents}, we show some examples from the three tasks, in which the entities are replaced with the special tags.  

In order to alleviate the out-of-vocabulary problem, BERT uses WordPiece embedding \cite{wu2016google} and each input word will be split into subwords from WordPiece vocabulary. In addition, following the original BERT input representation, we add the segment and position information of the tokens in the input sentence through the segment embedding and position embedding.

\begin{table*}
\centering
\begin{tabular}{c c c c | c c c | c c c }
\hline

\hline

\hline

\hline
\multirow{2}{4em}{Model}&\multicolumn{3}{c|}{PPI}&\multicolumn{3}{c|}{DDI} & \multicolumn{3}{c}{ChemProt}\\
\cline{2-10}
&  Precision & Recall & F1 Score & Precision & Recall & F1 Score & Precision & Recall & F1 Score \\
\hline
BioBERT &79.0 & 83.3 & 81.0 &77.4 & 79.4 & 78.4 & 74.6 & 76.8 & 75.7\\
BioBERT+Att\_LL &80.7 & \textbf{84.4} & 82.5 &\textbf{81.3} & 80.1 & \textbf{80.7} & \textbf{76.5} & \textbf{77.1} & \textbf{76.8}\\
\hline
\hline
BioBERT+LSTM\_LL* &79.6 & 82.0 & 80.8 &79.7 & 76.5 & 78.1 & 75.7 & 74.5 & 75.1\\
BioBERT+Att\_LL* &\textbf{82.3}&83.5& \textbf{82.8} & 79.7& 77.6& 78.6 & 76.4& 74.5& 75.4\\
\hline
\hline
BioBERT+LSTM\_CLS \cite{song2020utilizing} &79.0 & 82.3 & 80.6 &77.6 & \textbf{80.4} & 79.0 & 73.8 & 76.4 & 75.1\\
BioBERT+Att\_CLS \cite{song2020utilizing} &79.9 & 83.6 & 81.7 &78.6 & 77.8 & 78.2 & 73.2 & 75.3 & 74.2\\
\hline

\hline

\hline

\hline
\end{tabular}
\captionsetup{width=.85\textwidth}
\caption{Model performance without [CLS] token and utilizing [CLS] from intermediate layers. BioBERT+LSTM\_LL*/BioBERT+Att\_LL*: models of fine-tuning without [CLS] token; BioBERT+LSTM\_CLS: model of applying LSTM on the [CLS] tokens of intermediate layers; BioBERT+Att\_CLS: model of applying attention mechanism on the [CLS] tokens of intermediate layers.}
\label{tab:comp2}
\end{table*}

\begin{figure*}
     \begin{subfigure}[b]{0.9\textwidth}
         \centering
         \includegraphics[width=1.1\textwidth]{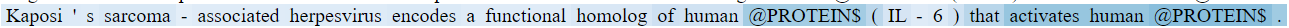}
         \caption{Example from PPI corpus (Label: Positive).}
         \label{fig:ppi}
     \end{subfigure}
     \vfill
     \begin{subfigure}[b]{0.9\textwidth}
         \centering
         \includegraphics[width=0.6\textwidth]{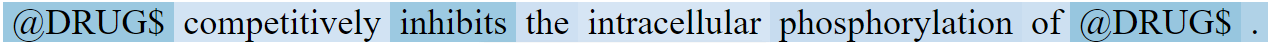}
         \caption{Example from DDI corpus (Label: EFFECT).}
         \label{fig:ddi}
     \end{subfigure}
     \vfill
     \begin{subfigure}[b]{0.9\textwidth}
         \centering
         \includegraphics[width=0.65\textwidth]{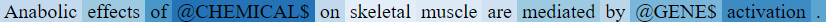}
         \caption{Example from ChemProt corpus (Label: CPR:3).}
         \label{fig:ddi}
     \end{subfigure}
        \caption{Attention weights visualization.}
        \label{fig:attention}
\end{figure*}
\section{Results and discussion}

In this section, we first discuss the model selection from the three BERT models for biomedical domain. Next, the results of new fine-tuning process are provided. At last, more discussions on attention weights are given to help us better understand the information summarization in attention mechanism.

As mentioned earlier, there are three adapted BERT models for biomedical domain: BioBERT, SciBERT and BlueBERT. The difference between the three models adapted to the biomedical domain is the text used for the pre-training. After fine-tuning on the corpora of the three tasks, BioBERT generally obtains good results (Please refer to Table \ref{tab:bert} in the Appendix for the details of the performance of the three models). Therefore, we will report our experimental results of new fine-tuning process based on BioBERT.
  
\subsection{Model performance after utilizing all the information from the last layer}

To provides context for the performance of the proposed method, we first report the state-of-the-art (SOTA) results and the performance of original BioBERT model on the three tasks in the first two rows of Table~\ref{tab:comp}. As shown in the table, the BioBERT model outperforms the state-of-the-art method on PPI task, but it is not up to par with the current SOTA methods on DDI and ChemProt tasks.

In the last two rows of Table~\ref{tab:comp}, the model performance of including the knowledge of the last layer is illustrated. Clearly, the use of information in the last layer leads to improvement for all three tasks. The model of utilizing attention mechanism on the last layer achieves the best performance on all the tasks, with 1.5\%, 2.3\% and 1.1\% F1 score improvement on PPI, DDI and ChemProt respectively. We also see that the use of attention mechanism outperforms the LSTM method on summarizing the information of last layer.   

Furthermore, we achieve the state-of-the-art performance on all three tasks after adding the attention mechanism on the last layer. Specifically, our method leads to 4.9\%, 0.8\% and 0.3\% F1 score improvement on PPI, DDI and ChemProt respectively. 

\subsection{Other fine-tuning mechanisms}

The results of new fine-tuning of BERT model that includes output information from both [CLS] token and the other tokens of the last layer have shown that this method achieves better performance than original fine-tuning mechanism. In this work, we also experiment with the method of fine-tuning without [CLS] output to explore the contribution of [CLS] token. In Table~\ref{tab:comp2}, the two rows in the middle (BioBERT+LSTM\_LL* and BioBERT+Att\_LL*) illustrate the model performance of utilizing the outputs of last layer without [CLS] token. The first two rows are only repetitions of the results from Table \ref{tab:comp} and provide the context for the new results. We can see that the fine-tuning without [CLS] achieves better performance with small margin on PPI task, but it lags behind the previous methods on the other two tasks. The results indicate that the [CLS] token contains key information for our classification tasks. Thus, utilizing all the outputs from the last layer generally gives us better results.    

In addition, we also compare our method with the mechanism of utilizing the [CLS] outputs from the intermediate layers. The last two rows in Table~\ref{tab:comp2} show the performance of BioBERT model after adding the information from the [CLS] outputs of all intermediate layers. As it is shown, our method of using information from the last layer achieves better performance than the mechanism of summarizing the [CLS] outputs from the intermediate layers. Compared with the original BioBERT model, we even observe the drop in the performance after including the [CLS] outputs from the intermediate layers in some cases (e.g., BioBERT+LSTM\_CLS on PPI task). Those results imply that [CLS] outputs in the intermediate layers cannot provide enough useful information for our relation extraction tasks.

\subsection{Visualization of attention weights}

The information acquired by attention mechanism from the last layer has been shown to improve the model performance. In this section, we will visualize the attention weight distribution on the words of sentences and examine the words the model is paying more attention to. 

We select an example from each of the three tasks that were misclassified by the original BioBERT model but are correctly predicted by the new method. In Fig \ref{fig:attention}, we visualize the weights based on the darkness of the color, with words in darker color having larger weights.  

We can see that the attention mechanism is giving large weights on the important words that indicate the relation between the entities. Typically, the word that is used to express the relation in a sentence is called "trigger word" of the relation. The attention mechanism is gaining the knowledge of "trigger word" in our tasks. For example, in the PPI sentence (Fig \ref{fig:attention}a), the words of entity token (@PROTEIN\$) and the word "activates" (trigger word of PPI) have larger weights than other words. Similarly, the trigger words "inhibit" and "mediate" have larger attention weights in the DDI and ChemProt examples respectively. Recall these sentences were not correctly predicted when attention was not used to highlight the information from the “trigger word", which also demonstrates the importance of the information from the last layer. 

\subsection{More analysis of attention weights}

After providing some examples of attention weight distribution in the sentences, it is useful to conduct an analysis on the attention weights and investigate the overall weight distribution on the words in the corpora.

Previously, we have shown that the attention mechanism will learn relatively larger weights on trigger word of the relations using three examples. In order to understand the attention weight especially on trigger words in a more general way, we need to consider all the words in the corpora. Given the fact that the trigger words usually appear near the entities of the relation, we collect all the weights of three words around the entities in the positive instances and take the average of the weights to acquire the global attention for each word. Considering the different forms of the words, we utilize Porter’s stemmer \cite{porter1980algorithm} to remove the morphological affixes from each word and only use the word stem for the global attention weight calculation. For instance, the stem for the word "activate" is "activ", and many other words like "activates" and "activation" have the same word stem. 

\begin{table}[h]
\centering
\begin{tabular}{c|c}
\hline

\hline

\hline

\hline
Task& Word Stem \\
 \hline
\multirow{2}{3.8em}{PPI}  & activ(ate), complex, associ(ate), interact, human\\
&protein, bind, domain, specif(y), receptor\\
 \hline
\multirow{2}{3.8em}{DDI} & concomitantli, combin(e), concomit(ant), increas(e), use \\ 
&concurr(ent), decreas(e), inhibit, receiv(e), administ(er)\\
\hline
\multirow{2}{3.8em}{ChemProt}& phosphoryl(ate), attenu(ate), stimul(ate), deriv(e), regul(ate) \\
&novel, metabol(ize), reduc(e), induc(e), inhibit  \\
\hline

\hline

\hline

\hline

\end{tabular}
\captionsetup{width=0.45\textwidth}
\caption{Top words with large attention weight from corpora of PPI, DDI and ChemProt tasks.}
\label{tab:word3}
\end{table}

In Table \ref{tab:word3}, we show the top 10 words that have large learned weight from the attention mechanism for the dataset of each task. Among the top words, we can see that most of them are "trigger words" of the relations, which indicates that the attention mechanism is learning the knowledge of relations from the last layer of BERT model.

\begin{figure}[tb]
\centering
\includegraphics[width=0.45\textwidth]{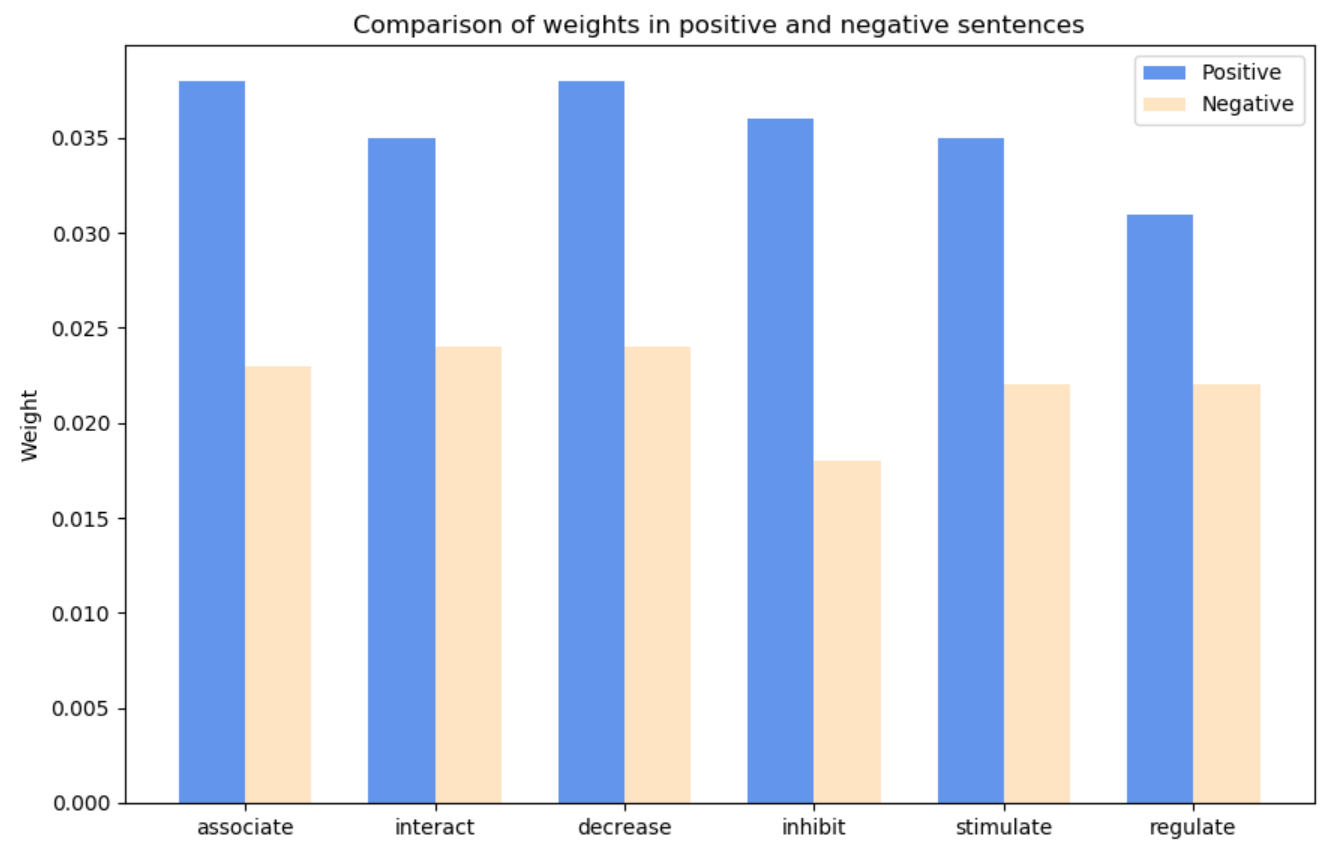}
\caption{\label{fig:weightcomp} Comparison of trigger word weights in positive and negative instances. The first two words (associate and interact) are from PPI task, the middle two words (decrease and inhibit) are from DDI task and the last two words (stimulate and regulate) are from ChemProt task.}
\end{figure}

\subsection{Trigger word weight comparison between positive and negative instances}

We have seen that the attention mechanism is giving relatively large weights on trigger words of relations on positive instances. For the negative instances, small weights should be learned on trigger words since there is no relationship between the entities. Therefore, we compare the trigger word weights between positive and negative instances and give six examples in Fig \ref{fig:weightcomp} from the three tasks. We can see that the learned trigger word weights in positive instances are much larger than the weights in the negative instances. This demonstrates that the attention mechanism can automatically learn the difference between positive and negative instances. 

\section{Conclusion}
In this paper, we propose a new fine-tune process for BERT model to utilize all the outputs from the last layer. We verify the effectiveness of the method on three relation extraction tasks: PPI, DDI and ChemProt. The experimental results illustrate that the proposed method achieves state-of-the-art performance on all three tasks. Also, the visualization and analysis of the weights in attention mechanism also demonstrates that our method is learning the key knowledge of the relations in the sentence to help the relation extraction. In the future, We will continue our investigation on extraction of other relations between biomedical entities. We will also investigate how our method can help other tasks in biomedical domain. 
\bibliographystyle{IEEEtran}
\bibliography{IEEEabrv,IEEEexample}

\appendix

\section*{Comparison of BERT models in biomedical domain}

We provide the performance of BioBERT, SciBERT and BlueBERT in biomedical domain on the PPI, DDI and ChemProt tasks. The best result on each metric is highlighted with bond font in Table \ref{tab:bert}.
\begin{table}[h]
\centering
\begin{tabular}{c c | c | c | c}
\hline

\hline

\hline

\hline
Task& Metrics&BioBERT&SciBERT& BlueBERT \\
 \hline
\multirow{3}{3em}{PPI}  & Precision& \textbf{79.0}&78.4&69.3\\
 & Recall& \textbf{83.3}&79.5&75.0\\
 & F score& \textbf{81.0}&78.8&71.9\\
 \hline
\multirow{3}{3em}{DDI}  & Precision& 77.4&\textbf{80.1}&77.7\\
 & Recall& \textbf{79.4}&78.1&75.4\\
 & F score& 78.4&\textbf{79.1}&76.5\\
\hline
\multirow{3}{3em}{ChemProt}  & Precision& \textbf{74.6}&74.3&70.8\\
 & Recall& \textbf{76.8}&73.2&64.8\\
 & F score& \textbf{75.7}&74.0&67.7\\

\hline

\hline

\hline

\hline

\end{tabular}
\captionsetup{width=0.45\textwidth}
\caption{Performance of three BERT models in biomedical domain on PPI, DDI and ChemProt tasks.}
\label{tab:bert}
\end{table}
\end{document}